\documentclass[10pt,twocolumn,letterpaper]{article}

\usepackage[preprint]{cvpr}      
\usepackage{graphicx}

\usepackage[utf8]{inputenc} 
\usepackage[T1]{fontenc}    
\usepackage{url}            
\usepackage{booktabs}       
\usepackage{amsfonts}       
\usepackage{nicefrac}       
\usepackage{microtype}      
\usepackage{xcolor}         
\usepackage{adjustbox}

\usepackage{bbding}
\usepackage{multirow}
\usepackage{booktabs}
\usepackage{makecell}
\usepackage{bbm}
\usepackage{courier}
\usepackage{caption}
\usepackage{float}

\usepackage[pagebackref,breaklinks,colorlinks]{hyperref}
\usepackage{multicol}

\usepackage[capitalize]{cleveref}
\crefname{section}{Sec.}{Secs.}
\Crefname{section}{Section}{Sections}
\Crefname{table}{Table}{Tables}
\crefname{table}{Tab.}{Tabs.}

\def\cvprPaperID{5306} 
\def\confName{CVPR}
\def\confYear{2023}

\begin{document}

\title{Mask DINO: Towards A Unified Transformer-based Framework for Object Detection and Segmentation}

\author{\textbf{Feng Li$^{1,3}$\thanks{Equal contribution.}\hspace{.4mm} \thanks{This work was done when Feng Li and Hao Zhang were interns at IDEA. }\hspace{1.5mm}, ~Hao Zhang$^{1,3*\dag}$, ~Huaizhe Xu$^{1,3}$, ~Shilong Liu$^{2,3}$},\\ \textbf{~Lei Zhang$^{3}$\thanks{Corresponding author.}\hspace{1.mm}, ~Lionel M. Ni$^{1,4}$, ~Heung-Yeung Shum$^{1,3}$} \\
$^1$The Hong Kong University of Science and Technology. \\
$^2$Dept. of CST., BNRist Center, Institute for AI, Tsinghua University. \\
$^3$International Digital Economy Academy (IDEA). \\
$^4$The Hong Kong University of Science and Technology (Guangzhou).\\
\texttt{\{fliay,hzhangcx,hxubr\}@connect.ust.hk} \\
\texttt{\{liusl20\}@mails.tsinghua.edu.cn} \\
\texttt{\{leizhang\}@idea.edu.cn} \\
\texttt{\{ni,hshum\}@ust.hk} \\
}
\maketitle

\begin{abstract}
In this paper we present Mask DINO, a unified object detection and segmentation framework. 
Mask DINO extends DINO (DETR with Improved Denoising Anchor Boxes) by adding a mask prediction branch which supports all image segmentation tasks (instance, panoptic, and semantic). 
It makes use of the query embeddings from DINO to dot-product a high-resolution pixel embedding map to predict a set of binary masks. 
Some key components in DINO are extended for segmentation through a shared architecture and training process. 
Mask DINO is simple, efficient, and scalable, and it can benefit from joint large-scale detection and segmentation datasets. 
Our experiments show that Mask DINO significantly outperforms all existing specialized segmentation methods, both on a ResNet-50 backbone and a pre-trained model with SwinL backbone. 
Notably, Mask DINO establishes the best results to date on instance segmentation (54.5 AP on COCO), panoptic segmentation (59.4 PQ on COCO), and semantic segmentation (60.8 mIoU on ADE20K) among models under one billion parameters. 
Code is available at \url{https://github.com/IDEA-Research/MaskDINO}.

\end{abstract}

\section{Introduction}
Object detection and image segmentation are fundamental tasks in computer vision. Both tasks are concerned with localizing objects of interest in an image but have different levels of focus. Object detection is to localize objects of interest and predict their bounding boxes and category labels, whereas image segmentation focuses on pixel-level grouping of different semantics. Moreover, image segmentation encompasses various tasks including instance segmentation, panoptic segmentation, and semantic segmentation with respect to different semantics, e.g., instance or category membership, foreground or background category. 

Remarkable progress has been achieved by classical convolution-based algorithms developed for these tasks with specialized architectures, such as Faster RCNN~\cite{ren2015faster} for object detection, Mask RCNN~\cite{he2017mask} for instance segmentation, and FCN~\cite{long2015fully} for semantic segmentation. Although these methods are conceptually simple and effective, they are tailored for specialized tasks and lack the generalization ability to address other tasks. The ambition to bridge different tasks gives rise to more advanced methods like HTC~\cite{chen2019hybrid} for object detection and instance segmentation and Panoptic FPN~\cite{kirillov2019panoptic}, K-net~\cite{zhang2021k} for instance, panoptic, and semantic segmentation. 
Task unification not only helps simplify algorithm development but also brings in performance improvement in multiple tasks. 

Recently, DETR-like \cite{carion2020end} models developed based on Transformers~\cite{vaswani2017attention} have achieved inspiring progress on many detection and segmentation tasks.
As an end-to-end object detector, DETR adopts a set-prediction objective and eliminates hand-crafted modules such as anchor design and non-maximum suppression.
Although DETR addresses both the object detection and panoptic segmentation tasks, its segmentation performance is still inferior to classical segmentation models. To improve the detection and segmentation performance of Transformer-based models, researchers have developed specialized models for object detection ~\cite{zhu2020deformable,liu2022dab, li2022dn, zhang2022dino}, image segmentation~\cite{zhang2021k,cheng2021maskformer, cheng2021mask2former}, instance segmentation~\cite{fang2021instances}, panoptic segmentation~\cite{qin2022pyramid}, and semantic segmentation~\cite{jain2021semask}.

Among the efforts to improve object detection, DINO~\cite{zhang2022dino} takes advantage of the dynamic anchor box formulation from DAB-DETR~\cite{liu2022dab} and query denoising training from DN-DETR~\cite{li2022dn}, and further 
achieves the SOTA result on the COCO object detection leaderboard for the first time as a DETR-like model. Similarly, for improving image segmentation, MaskFormer~\cite{cheng2021maskformer} and Mask2Former~\cite{cheng2021mask2former} propose to unify different image segmentation tasks using query-based Transformer architectures to perform mask classification. Such methods have achieved remarkable performance improvement on multiple segmentation tasks.

However, in Transformer-based models, the best-performing detection and segmentation models are still not unified, which 
prevents task and data cooperation between detection and segmentation tasks.
As an evidence, in CNN-based models, Mask-R-CNN~\cite{he2017mask} and HTC~\cite{chen2019hybrid} are still widely acknowledged as unified models that achieve mutual cooperation between detection and segmentation to achieve superior performance than specialized models.
Though we believe detection and segmentation can help each other in a unified architecture in Transformer-based models, the results of simply using DINO for segmentation and using Mask2Former for detection indicate that they can not do other tasks well, as shown in Table \ref{tab:mask2former_head} and \ref{tab:dino_head}. Moreover, trivial multi-task training can even hurt the performance of the original tasks. It naturally leads to two questions: 1) \emph{why cannot detection and segmentation tasks help each other in Transformer-based models?} and 2) \emph{is it possible to develop a unified architecture 
to replace specialized ones?}

To address these problems, we propose Mask DINO, which extends DINO with a mask prediction branch in parallel with DINO's box prediction branch. Inspired by other unified models~\cite{wang2021max,cheng2021maskformer,cheng2021mask2former} for image segmentation, we reuse content query embeddings from DINO to perform mask classification for all segmentation tasks on a high-resolution pixel embedding map (1/4 of the input image resolution) obtained from the backbone and Transformer encoder features. 
The mask branch predicts binary masks by simply dot-producting each content query embedding with the pixel embedding map. 
As DINO is a detection model for region-level regression, it is not designed for pixel-level alignment.
To better align features between detection and segmentation, we also propose three key components to boost the segmentation performance.
First, we propose a unified and enhanced query selection. It utilizes encoder dense prior by predicting masks from the top-ranked tokens to initialize mask queries as anchors. In addition, we observe that pixel-level segmentation is easier to learn in the early stage and propose to use initial masks to enhance boxes, which achieves task cooperation. Second, we propose a unified denoising training for masks to accelerate segmentation training. 
Third, we use a hybrid bipartite matching for more accurate and consistent matching from ground truth to both boxes and masks.

Mask DINO is conceptually simple and easy to implement under the DINO framework. 
To summarize, our contributions are three-fold.
    \textbf{1)} We develop a unified Transformer-based framework for both object detection and segmentation. As the framework is extended from DINO, by adding a mask prediction branch, it naturally inherits most algorithm improvements in DINO including anchor box-guided cross attention, query selection, denoising training, and even a better representation pre-trained on a large-scale detection dataset.
    \textbf{2)} We demonstrate that detection and segmentation can help each other through a shared architecture design and training method. Especially, detection can significantly help segmentation tasks, even for segmenting background "stuff" categories.
    Under the same setting with a ResNet-50 backbone, Mask DINO outperforms all existing models compared to DINO ($+0.8$ AP on COCO detection) and Mask2Former ($+\textbf{2.6}$ AP, $+\textbf{1.1}$ PQ, and $+\textbf{1.5}$ mIoU on COCO instance, COCO panoptic, and ADE20K semantic segmentation).
    \textbf{3)} We also show that, via a unified framework, segmentation can benefit from detection pre-training on a large-scale detection dataset. After detection pre-training on the Objects365~\cite{shao2019objects365} dataset with a SwinL~\cite{liu2021swin} backbone, Mask DINO significantly improves all segmentation tasks and achieves the best results on instance (\textbf{54.5} AP on COCO),  panoptic (\textbf{59.4} PQ on COCO), and semantic (\textbf{60.8} mIoU on ADE20K) segmentation among models under one billion parameters. 
\section{Related Work}
\noindent\textbf{Detection:} 
Mainstream detection algorithms have been dominated by convolutional neural network-based frameworks, until recently Transformer-based detectors~\cite{carion2020end,liu2022dab,li2022dn,zhang2022dino} achieve great progress. DETR~\cite{carion2020end} is the first end-to-end and query-based Transformer object detector, which adopts a set-prediction objective with bipartite matching. DAB-DETR~\cite{liu2022dab} improves DETR by formulating queries as $4$D anchor boxes and refining predictions layer by layer. DN-DETR~\cite{li2022dn} introduces a denoising training method to accelerate convergence. 
Based on DAB-DETR and DN-DETR, DINO~\cite{zhang2022dino} proposes several new improvements on denoising and anchor refinement and achieves new SOTA results on COCO detection. Despite the inspiring progress, DETR-like detection models are not competitive for segmentation. Vanilla DETR incorporates a segmentation head in its architecture. However, its segmentation performance is inferior to specialized segmentation models and only shows the feasibility of DETR-like detection models to deal with detection and segmentation simultaneously.
\\\textbf{Segmentation:}
Segmentation mainly includes instance, semantic, and panoptic segmentation. 
Instance segmentation is to predict a mask and its corresponding category for each object instance. Semantic segmentation requires to classify each pixel including the background into different semantic categories. Panoptic segmentation~\cite{kirillov2019panoptic} unifies the instance and semantic segmentation tasks and predicts a mask for each object instance or background segment. In the past few years, researchers have developed specialized architectures for the three tasks. For example, Mask-RCNN~\cite{he2017mask} and HTC~\cite{chen2019hybrid} can only deal with instance segmentation because they predict the mask of each instance based on its box prediction. FCN~\cite{long2015fully} and U-Net~\cite{ronneberger2015u} can only perform semantic segmentation since they predict one segmentation map based on pixel-wise classification. Although models for panoptic segmentation~\cite{kirillov2019pafpn,xiong2019upsnet} unifies the above two tasks, they are usually inferior to specialized instance and semantic segmentation models. Until recently, some  image segmentation models~\cite{zhang2021k,cheng2021maskformer, cheng2021mask2former} are developed to unify the three tasks with a universal architecture. For instance, Mask2Former~\cite{cheng2021mask2former} improves MaskFormer~\cite{cheng2021maskformer} by introducing masked-attention to Transformer.
Mask2Former has a similar architecture as DETR to probe image features with learnable queries but differs in using a different segmentation branch and some specialized designs for mask prediction. However, while Mask2Former shows a great success in unifying all segmentation tasks, it leaves object detection untouched and our empirical study shows that its specialized architecture design is not suitable for predicting boxes.
\\\textbf{Unified Methods:}
As both object detection and segmentation are concerned with localizing objects, they naturally share common model architectures and visual representations. A unified framework not only helps simplify the algorithm development effort, but also allows to use both detection and segmentation data to improve representation learning. There have been several previous works to unify segmentation and detection tasks, e.g., Mask RCNN~\cite{he2017mask}, HTC~\cite{chen2019hybrid}, and DETR~\cite{carion2020end}. Mask RCNN extends Faster RCNN and pools image features from Region Of Interest (ROI) proposed by RPN.
HTC further proposes an interleaved way of predicting boxes and masks. 
However, these two models can only perform instance segmentation. DETR predicts boxes and masks together in an end-to-end manner. However, its segmentation performance largely lags behind other models. According to Table~\ref{tab:dino_head}, adding DETR's segmentation head to DINO results in inferior instance segmentation results.
How to attain mutual assistance between segmentation and detection 
has long been an important problem to solve.

\section{Mask DINO}
Mask DINO is an extension of DINO~\cite{zhang2022dino}. On top of content query embeddings, DINO has two branches for box prediction and label prediction. The boxes are dynamically updated and used to guide the deformable attention in each Transformer decoder. Mask DINO adds another branch for mask prediction and minimally extends several key components in detection to fit segmentation tasks. To better understand Mask DINO, we start by briefly reviewing DINO and then introduce Mask DINO.
\subsection{Preliminaries: DINO}
\begin{figure*}[h]
    \includegraphics[width=0.9\textwidth]{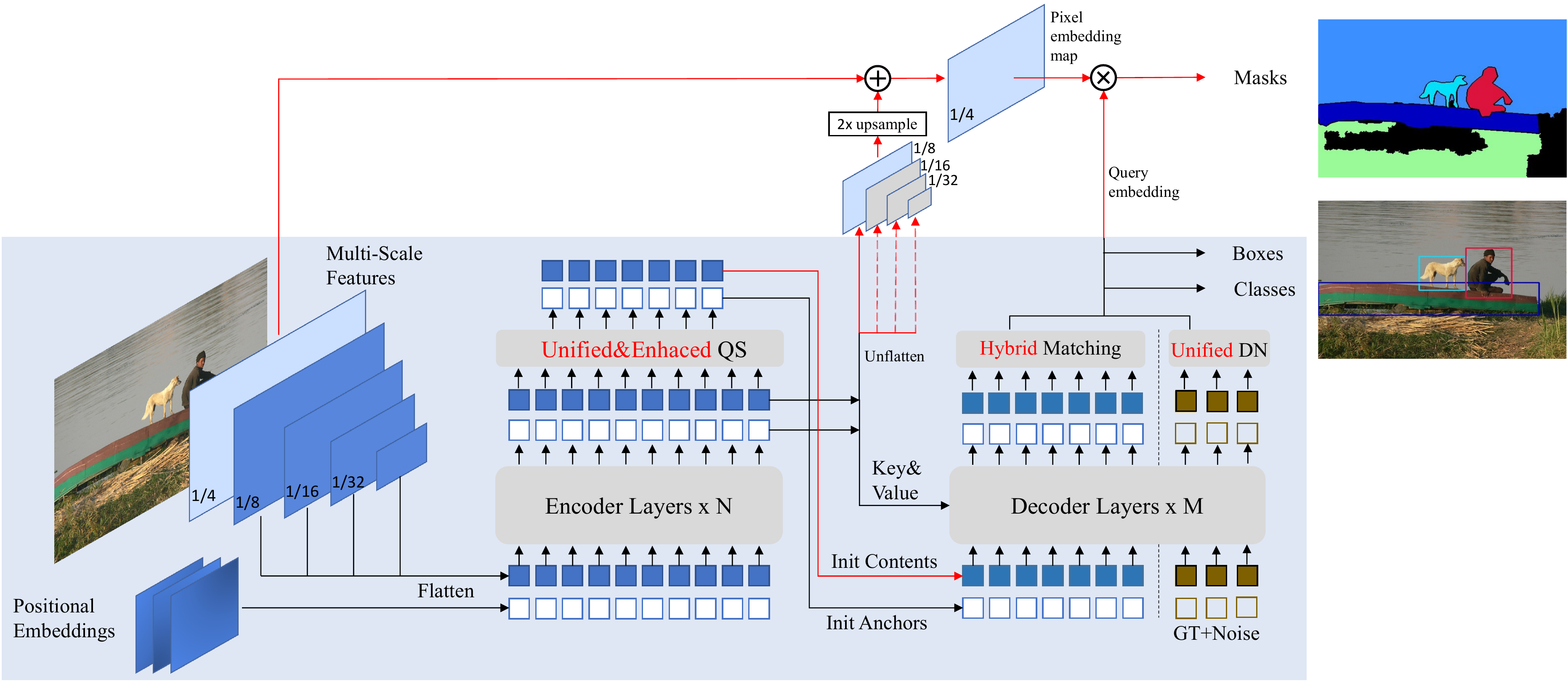}
    \centering
    \caption{The framework of Mask DINO, which is based on DINO (the blue-shaded part) with extensions (the red part) for segmentation tasks. 'QS' and 'DN' are short for query selection and denoising training, respectively. 
    }
    \label{fig:framework}
\end{figure*}

DINO is a typical DETR-like model, which is composed of a backbone, a Transformer encoder, and a Transformer decoder. The framework is shown in Fig.~\ref{fig:framework} (the blue-shaded part without red lines). Following DAB-DETR~\cite{liu2022dab}, DINO formulates each positional query in DETR as a 4D anchor box, which is dynamically updated through each decoder layer. Note that DINO uses multi-scale features with deformable attention~\cite{zhu2020deformable}. Therefore, the updated anchor boxes are also used to constrain deformable attention in a sparse and soft way. Following DN-DETR~\cite{li2022dn}, DINO adopts denoising training and further develops contrastive denoising to accelerate training convergence. Moreover, DINO proposes a mixed query selection scheme to initialize positional queries in the decoder and a look-forward-twice method to improve box gradient back-propagation.
\begin{table*}[t]
\begin{adjustbox}{width=0.99\textwidth,center}
\begin{minipage}[t]{0.55\textwidth}
\makeatletter\def\@captype{table}
\centering
\begin{tabular}{c|ccc}
    \toprule
    Model & Box AP & Mask AP  \\
        \midrule
        Mask2Former & ${46.2}^*$ & $43.7$     \\
        \midrule
        Mask2Former + detection head&${21.6}$&$41.3$\\
        \midrule
     DINO     &$\textbf{50.7}$    & $-$  \\
    
    \bottomrule
\end{tabular}
\caption{Simply adding a detection head to Mask2Former results in low detection performance. $^*$ indicates the boxes are generated from the predicted masks. The generated boxes from Mask2Former are also inferior to DINO (-4.5 AP). The models are trained for 50 epochs.
}
\label{tab:mask2former_head}
\end{minipage}\hspace{3mm}

\begin{minipage}[t]{0.65\textwidth}
\makeatletter\def\@captype{table}
\centering
\begin{tabular}{c|ccc}
    \toprule
    Model & Box AP & Mask AP  \\
        \midrule
        DINO + Mask2Former segmentation head&${49.9}$&$40.2$\\
        \midrule
     \begin{tabular}[c]{@{}c@{}}DINO + DETR segmentation head \\(finetune DINO pretrained on COCO detection)\end{tabular}      &${-}$    & $35.8$  \\
     \midrule
     Mask2Former &$-$ &$\textbf{43.7}$ \\
    \bottomrule
\end{tabular}
\caption{Simply adding a segmentation head to DINO results in low instance segmentation performance.  The predicted masks from DINO are also inferior to Mask2Former (-3.5 AP). The models are trained for 50 epochs.
}
\label{tab:dino_head}
\end{minipage}\hspace{3mm}

\end{adjustbox}
\vspace{-.4cm}
\end{table*}

\subsection{Why a universal model has not replaced the specialized models in DETR-like models?}\label{sec:why_spec}
Remarkable progress has been achieved by Transformer-based detectors and segmentation models. For instance, DINO~\cite{zhang2022dino} and Mask2Former~\cite{cheng2021mask2former} have achieved the best results on COCO detection and panoptic segmentation, respectively.
Inspired by such progress, we attempted to simply extend these specialized models for other tasks but found that the performance of other tasks lagged behind the original ones by a large margin, as shown in Table \ref{tab:mask2former_head} and \ref{tab:dino_head}. It seems that trivial multi-task training even hurts the performance of the original task. However, in convolution-based models, it has shown effective and mutually beneficial to combine detection and instance segmentation tasks. For example, detection models with Mask R-CNN head~\cite{he2017mask} is still ranked the first on the COCO instance segmentation. 
We will take DINO and Mask2Former as examples to discuss the challenges in unifying Transformer-based detection and segmentation.
\\\textbf{$-$What are the differences between specialized detection and segmentation models?}
Image segmentation is a pixel-level classification task, while object detection is a region-level regression task. In DETR-based model, the decoder queries are responsible for these tasks. For example, Mask2Former uses such decoder queries to dot-product the high-resolution feature maps to produce segmentation masks, while DINO uses them to regress boxes. However, as such queries in Mask2Former only have to compare per-pixel similarity with the image features, they may not be aware of the region-level position of each instance. On the contrary, queries in DINO are not designed to interact with such low-level features to learn pixel-level representation. Instead, they encode rich positional information and high-level semantics for detection.
\\\textbf{$-$Why cannot Mask2Former do detection well?}
The Transformer decoder of Mask2Former is designed for segmentation tasks and does not suit detection for three reasons. First, its \textbf{queries} follow the design in DETR~\cite{carion2020end} without being able to utilize better positional priors as studied in Conditional DETR~\cite{meng2021conditional}, Anchor DETR~\cite{wang2021anchor}, and DAB-DETR~\cite{liu2022dab}. For example, its content queries are semantically aligned with the features from the Transformer encoder, whereas its positional queries are just learnable vectors as in vanilla DETR instead of being associated with a single-mode position~\footnote{We refer the interested readers to  discussions in Sec. 3 in DAB-DETR~\cite{liu2022dab}}. If we remove its mask branch, it reduces to a variant of DETR~\cite{carion2020end}, whose performance is inferior to recently improved DETR models.
Second, Mask2Former adopts \textbf{masked attention} (multi-head attention with attention mask) in Transformer decoders. The attention masks predicted from a previous layer are of high resolution and used as hard-constraints for attention computation. They are neither efficient nor flexible for box prediction. Third, Mask2Former cannot explicitly perform \textbf{box refinement} layer by layer. Moreover, its coarse-to-fine mask refinement in decoders fails to use multi-scale features from the encoder. As shown in Table~\ref{tab:mask2former_head}, the generated box AP from mask is 4.5 AP worse than DINO and trivial multi-task learning by adding a detection head is not working~\footnote{We also notice there are issues in official Mask2Former Github  (https://github.com/facebookresearch/Mask2Former/issues/43) that fail to make Mask2Former work well by adding a detection head.}.
\\\textbf{$-$Why cannot DETR/DINO do segmentation well?}
As shown in Table~\ref{tab:dino_head}, simply \textbf{1)} adding DETR's segmentation head or \textbf{2)} adding Mask2Former's segmentation head result in inferior performance compared to Mask2Former.
We analyze the reasons as follows.  \textbf{The reason for 1)} is that DETR's {segmentation head} is not optimal. The vanilla DETR lets each query embedding dot-product with the smallest feature map to compute attention maps and then upsamples them to get the mask predictions. This design lacks an interaction between queries and larger feature maps from the backbone. In addition, the head is too heavy to use mask auxiliary loss for mask refinement.
\textbf{The reason for 2)} is that features in improved detection models are not aligned with segmentation. For example, DINO inherits many designs from \cite{zhu2020deformable, liu2022dab, zhang2022dino} like query formulation, denoising training, and query selection. However, these components are designed to strengthen region-level representation for detection, which is not optimal for segmentation.

\subsection{Our Method: Mask DINO}
Mask DINO adopts the same architecture design for detection as in DINO with minimal modifications. In the Transformer decoder, Mask DINO adds a mask branch for segmentation and extends several key components in DINO for segmentation tasks. As shown in Fig.~\ref{fig:framework}, the framework in the blue-shaded part is the original DINO model and the additional design for segmentation is marked with red lines. 
\subsection{\textbf{Segmentation branch}} Following other unified models~\cite{wang2021max,cheng2021maskformer,cheng2021mask2former} for image segmentation, we perform mask classification for all segmentation tasks. Note that DINO is not designed for pixel-level alignment as its positional queries are formulated as anchor boxes and its content queries are used to predict box offset and class membership. To perform mask classification, we adopt a key idea from Mask2Former~\cite{cheng2021mask2former} to construct a pixel embedding map which is  obtained from the backbone and Transformer encoder features. As shown in Fig.  ~\ref{fig:framework}, the pixel embedding map is obtained by fusing the $1/4$ resolution feature map $C_b$ from the backbone with an upsampled $1/8$ resolution feature map $C_e$ from the Transformer encoder. Then we dot-product each content query embedding $q_c$ from the decoder with the pixel embedding map to obtain an output mask $m$. 
\begin{equation}
    m = q_{c} \otimes \mathcal{M}(\mathcal{T}(C_{b})+\mathcal{F}(C_{e})),
\end{equation}
where $\mathcal{M}$ is the segmentation head, $\mathcal{T}$ is a convolutional layer to map the channel dimension to the Transformer hidden dimension, and $\mathcal{F}$ is a simple interpolation function to perform 2x upsampling of $C_e$. This segmentation branch is conceptually simple and easy to implement in the DINO framework, as shown in Fig.~\ref{fig:framework}.
\subsection{Unified and Enhanced Query Selection}
\noindent\textbf{Unified query selection for mask:} 
Query selection has been widely used in traditional two-stage models \cite{ren2015faster} and many DETR-like models \cite{zhu2020deformable, zhang2022dino} to improve detection performance.
We further improve the query selection scheme in Mask DINO for segmentation tasks. 

The encoder output features contain dense features, which can serve as better priors for the decoder. Therefore, we adopt three prediction heads (classification, detection, and segmentation) in the encoder output. Note that the three heads are identical to the decoder heads. The classification score of each token is considered as the confidence to select top-ranked features and feed them to the decoder as content queries. The selected features also regress boxes and dot-product with the high-resolution feature map to predict masks. The predicted boxes and masks will be supervised by the ground truth and are considered as initial anchors for the decoder.
Note that we initialize both the content and anchor box queries in Mask DINO whereas DINO only initializes anchor box queries.
\\\textbf{Mask-enhanced anchor box initialization: }As summarized in Sec \ref{sec:why_spec}, image segmentation is a pixel-level classification task while object detection is a region-level position regression task. Therefore, compared to detection, though segmentation is a more difficult task with fine-granularity, it is easier to learn in the initial stage. For example, masks are predicted by dot-producting queries with the high-resolution feature map, which only needs to compare per-pixel semantic similarity. However, detection requires to directly regress the box coordinates in an image. Therefore, in the initial stage after unified query selection, mask prediction is much more accurate than box (the qualitative AP comparison between mask prediction and box prediction in different stages is also shown in Table \ref{tab:dec_test} and \ref{tab:maskenhance}). Therefore, after unified query selection, we derive boxes from the predicted masks as better anchor box initialization for the decoder. By this effective task cooperation, the enhanced box initialization can bring in a large improvement to the detection performance.
\begin{table*}[t]
    \centering
    \begin{adjustbox}{width=0.99\textwidth,center}
    \begin{tabular}{l|c|c|cc|ccccc|c|c|c}
    \toprule
        Model  & Epochs & Query type &Mask AP &Box AP & AP$_{50}^{mask}$ & AP$_{75}^{mask}$ & AP$_{S}^{mask}$ & AP$_{M}^{mask}$ & AP$_{L}^{mask}$ & GFLOPS & Params&FPS \\
        \midrule
        \textbf{ResNet-50 backbone}\\
        \toprule
        Mask-RCNN~\cite{he2017mask,du2021simple,ghiasi2021simple} & $400$ & Dense anchors  & $42.5$ & $48.2$ & $-$ & $-$ & $23.8$ & $45.0$ & $60.0$ & $207$ & $40$M & $10.3$ \\
        HTC \cite{chen2019hybrid}& $36$ & Dense anchors & $39.7$ & $44.9$ & $61.4$ & $43.1$ & $22.6$ & $42.2$ & $50.6$ &$441$ & $80$M & $5$ \\
        QueryInst~\cite{fang2021instances} &36&300 queries&$40.6$&$45.6$&$63.0$&$44.0$&$23.4$&$42.5$&$52.8$&$-$&$-$&$7.0$ \\
        \hline
        DINO-4scale~\cite{zhang2022dino}    &36 & 900 queries & $-$ &\underline{50.9}& $-$ & $-$ & $-$ & $-$ & $-$&$245$ & $47$M & $19.6$\\
         \hline
        Mask2Former~\cite{cheng2021mask2former}    &12 & 100 queries& {38.7} & $-$& $59.8$ & $41.2$ & $18.2$ & $41.5$ & $59.8$ &$226$ & $44$M & $8.2$ \\
        Mask DINO (ours)    &12 & 300 queries  &\textbf{41.4}\fontsize{7.0pt}{\baselineskip}\selectfont{(+2.7)} &{45.7} & $62.9$ & $44.6$ & $21.1$ & $44.2$ & $61.4$ &$286$ & $52$M & $14.8$ \\
        \hline
         Mask DINO (ours)    &\textbf{24} & 300 queries  &\textbf{44.2}\fontsize{7.0pt}{\baselineskip}\selectfont{(+0.5)} &{48.4} & $66.6$ & $47.9$ & $23.9$ & $47.0$ & $64.0$ &$286$ & $52$M & $14.8$ \\
        \hline
        Mask2Former$^*$~\cite{cheng2021mask2former}    &50 & 100 queries& \underline{43.7} & 46.2$^{\dag}$ & $66.0$ & $46.9$ & $23.4$ & $47.2$ & $64.8$ &$226$ & $44$M & 8.2 \\
        Mask DINO (ours)    &50 & 100 queries  &{45.4} &{49.8} & $67.9$ & $49.3$ & $25.2$ & $48.3$ & $65.8$ &$280$ & $52$M & $15.2$ \\
        \hline
        Mask DINO (ours)    &50 & 300 queries 
        &\textbf{46.0}&{50.5} & $68.9$ & $50.3$ & \textbf{26.0}\fontsize{7.0pt}{\baselineskip}\selectfont{(+2.6)} & $49.3$\fontsize{7.0pt}{\baselineskip}\selectfont{(+2.1)} & $65.5$\fontsize{7.0pt}{\baselineskip}\selectfont{(+0.7)} &$286$ & $52$M & $14.8$ \\
        \begin{tabular}[l]{@{}l@{}}Mask DINO$^\ddag$ (ours)  \end{tabular}     &50 & 300 queries  &\textbf{46.3}\fontsize{7.0pt}{\baselineskip}\selectfont{(+2.6)} &\textbf{51.7}\fontsize{7.0pt}{\baselineskip}\selectfont{(+0.8)} & $69.0$ & $50.7$ & \textbf{26.1}\fontsize{7.0pt}{\baselineskip}\selectfont{(+2.7)} & $49.3$\fontsize{7.0pt}{\baselineskip}\selectfont{(+2.1)} & $66.1$\fontsize{7.0pt}{\baselineskip}\selectfont{(+1.3)} &$286$ & $52$M & $14.2$ \\
        \hline
        \textbf{SwinL backbone}\\
        \toprule
        HTC++ \cite{chen2019hybrid, liu2021swin}    &72 & dense anchors 
        &{$49.5$} &{57.1}& $-$ & $-$ & $31.0$ & $52.4$ & {67.2}  &$1470$ & $284$M & $-$ \\
        Mask2Former \cite{cheng2021mask2former}     &100 & 200 queries  &\underline{50.1} &{$-$} & $-$ & $-$ & {29.9} & $53.9$ & $72.1$ &868 & $216$M & $4.0$ \\
        DINO \cite{zhang2022dino}    &36 & 900 queries 
        &{$-$} &\underline{58.5} & $-$ & $-$ & {-} & $-$ & $-$ &$1285$ & $217$M & $8.1$ \\
        Mask2Former     &100 & 200 queries  &{50.1} &{$-$} & $-$ & $-$ & {29.9} & $53.9$ & $72.1$ &868 & $216$M & $4.0$ \\
        Mask DINO (ours)     &50 & 300 queries  &\textbf{52.1} &{58.3} & $76.5$ & $57.6$ & {32.9} & $55.4$ & $72.5$ &$1326$ & $223$M & $6.1$ \\
        Mask DINO$^\ddag$ (ours)     &50 & 300 queries  &\textbf{52.3}\fontsize{7.0pt}{\baselineskip}\selectfont{(+2.2)} &\textbf{59.0}\fontsize{7.0pt}{\baselineskip}\selectfont{(+0.5)} & $76.6$ & $57.8$ & {33.1} & $55.4$ & $72.6$ &$1326$ & $223$M & $5.6$ \\
        
        \bottomrule
    \end{tabular}
    \end{adjustbox}
    \centering
    \caption{Results for Mask DINO and other object detection and instance segmentation models with ResNet-50 and SwinL backbone on COCO \texttt{val2017} without extra data or tricks. Following DINO~\cite{zhang2022dino}, we use ResNet-50 with four feature scales by default, and use five scales under large models with a SwinL backbone. We follow the common practice in DETR-like models to use 300 queries. $^*$ Mask2Former using 300 queries is not listed as its performance will degenerate when using 300 queries. $^\dag$ indicates the box AP is derived from mask prediction. $^\ddag$ we use the proposed mask-enhanced box initialization to further improve detection performance. We test the FPS and GFLOPS of Mask2Former and Mask DINO on the A100 GPU using detectron2.
    }
    
    \label{tab:instance}
\end{table*}
\begin{table*}[t]
    \centering
    \begin{adjustbox}{width=0.8\textwidth,center}
    \begin{tabular}{l|c|c|ccc|ccc}
    \toprule
        Model  & Epochs & Query type &PQ &PQ$^{Th}$ & PQ$^{St}$ & Box AP$^{Th}_{pan}$ & Mask AP$^{Th}_{pan}$   \\
        \hline
        \textbf{ResNet-50 backbone}\\
        \midrule
        DETR~\cite{carion2020end} & $500+25$ & 100 queries  & $43.4$ & $48.2$ & $36$ & $-$ & $31.1$\\
        Panoptic Segformer~\cite{li2021panoptic} & 24& 353 queries&$49.6$&$54.4$&$42.4$& $-$&$41.7$\\
        Mask2Former$^*$ \cite{cheng2021mask2former}   &50 & 100 queries& {51.9}/$51.5^\dag$ & 57.7 & $43.0$ & $-$ & $41.7$ \\
        
        Mask DINO (ours)    &{50} & 100 queries  & $52.3$&$58.3$&$43.2$&$47.7$&$43.7$ \\
        Mask DINO (ours)    &50 & 300 queries  &\textbf{53.0}\fontsize{7.0pt}{\baselineskip}\selectfont{(+1.1)} &\textbf{59.1}\fontsize{7.0pt}{\baselineskip}\selectfont{(+1.4)} & $\textbf{43.9}$\fontsize{7.0pt}{\baselineskip}\selectfont{(+0.9)} & $\textbf{48.8}$ & \textbf{44.3}\fontsize{7.0pt}{\baselineskip}\selectfont{(+2.6)} \\
        Mask DINO (ours)    &\textbf{24} & 300 queries  & $51.5$ &$57.3$&$42.6$&$46.4$&$42.8$\\
         \hline        
        Mask2Former~\cite{cheng2021mask2former}    &12 & 100 queries& {46.9} & $52.5$ & $38.4$ & $-$ &  37.2\\
        Panoptic Segformer~\cite{li2021panoptic}&12&353 queries &$48.0$ & $52.3$ &$41.5$& $-$& $-$\\
        Mask DINO (ours)    &12 & 300 queries  &\textbf{49.0}\fontsize{7.0pt}{\baselineskip}\selectfont{(+1.0)} &{54.8} & $40.2$ & $43.2$ & $\textbf{40.4}$\fontsize{7.0pt}{\baselineskip}\selectfont{(+3.2)} \\
        \hline
        \textbf{SwinL backbone}\\
        \midrule
        Mask2Former~\cite{cheng2021mask2former}    &100 & 100 queries& {57.8} & $64.2$ & $48.1$ & $-$ &  48.6\\
        OneFormer~\cite{jain2022oneformer}&100&150 queries &$57.9$ & $64.4$ &$48.0$& $-$& $49.0$\\
        Mask DINO (ours)    &50 & 300 queries  &\textbf{58.3}\fontsize{7.0pt}{\baselineskip}\selectfont{(+0.5)} &{65.1} & $48.0$ & $56.2$ & $\textbf{50.6}$\fontsize{7.0pt}{\baselineskip}\selectfont{(+2.0)} \\
        \bottomrule
    \end{tabular}
    \end{adjustbox}
    \centering
    \caption{Results for Mask DINO and other panoptic segmentation models with a ResNet-50 backbone on COCO \texttt{val2017}. $^*$ Mask2Former using 300 queries is not listed as its performance will degenerate
when using 300 queries. 
$^\dag$ Our reproduced result.
    }
    
    \label{tab:panoptic}
    \vspace{-.4cm}
\end{table*}
\subsection{Segmentation Micro Design}
\noindent\textbf{Unified denoising for mask:} Query denoising in object detection has shown effective ~\cite{zhang2022dino,li2022dn} to accelerate convergence and improve performance. It adds noises to ground-truth boxes and labels and feed them to the Transformer decoder as noised positional queries and content queries. The model is trained to reconstruct ground truth objects given their noised versions. We also extend this technique to segmentation tasks. As masks can be viewed as a more fine-grained representation of boxes, box and mask are naturally connected. Therefore, we can treat boxes as a noised version of masks, and train the model to predict masks given boxes as a denoising task. The given boxes for mask prediction are also randomly noised for more efficient mask denoising training. The detailed noise and its hyperparameters used in our model are shown in Appendix \ref{sec:denoise}.
\\\textbf{Hybrid matching:} Mask DINO, as in some traditional models~\cite{chen2019hybrid, he2017mask}, predicts boxes and masks with two parallel heads in a loosely coupled manner. Hence the two heads can predict a pair of box and mask that are inconsistent with each other. To address this issue, in addition to the original box and classification loss in bipartite matching, we add a mask prediction loss to encourage more accurate and consistent matching results for one query. Therefore, the matching cost becomes $\lambda_{cls}\mathcal{L}_{cls}+\lambda_{box}\mathcal{L}_{box}+\lambda_{mask}\mathcal{L}_{mask}$, where $\mathcal{L}_{cls}, \mathcal{L}_{box}$, and $\mathcal{L}_{mask}$ are the classification, box, and mask loss and $\lambda$ are their corresponding weights. The detailed losses used in our model and their corresponding weights are shown in Appendix \ref{sec:generalimpl}.
\\\textbf{Decoupled box prediction:} For the panoptic segmentation task, box prediction for "stuff" categories is unnecessary and intuitively inefficient. For example, many "stuff" categories are background like "sky", whose GT mask-derived boxes are highly irregular and often cover the whole image. Therefore, box prediction for these categories can mislead the instance-level ("thing") detection and segmentation. To address this problem, we remove box loss and box matching for "stuff" categories. More specifically, the box prediction pipeline remains the same for "stuff" to locate meaningful regions and extract features with deformable attention. However, we do not count their box prediction loss. In our hybrid matching, the box loss for "stuff" is set to the mean of "thing" categories. This decoupled design can accelerate training and yield additional gains for panoptic segmentation.

\begin{table*}[t]
\begin{adjustbox}{width=0.9\textwidth,center}
\begin{minipage}[t]{0.7\textwidth}
\makeatletter\def\@captype{table}
\centering
\large
\begin{tabular}{c|cc|ccc}
    \toprule
        Model  & Iterations& \begin{tabular}[c]{@{}c@{}}Crop\\ size\end{tabular}&\begin{tabular}[c]{@{}c@{}}mIoU\\(mean)\end{tabular} &\begin{tabular}[c]{@{}c@{}}mIoU\\(high)\end{tabular}&\begin{tabular}[c]{@{}c@{}}mIoU\\(reported)\end{tabular}  \\
        \midrule
        Mask2Former~\cite{cheng2021mask2former} &160k& $512$&$46.1$ & $46.5$ &$47.2$  \\
          Mask DINO (ours)    & 160k& $512$&$\textbf{47.7}$\fontsize{7.0pt}{\baselineskip}\selectfont{(+1.6)} &\textbf{48.7}\fontsize{7.0pt}{\baselineskip}\selectfont{(+2.2)}&\textbf{48.7}\fontsize{7.0pt}{\baselineskip}\selectfont{(+1.6)}\\  \bottomrule
    \end{tabular}
    \centering
    \captionsetup{font={large}}
    \caption{{Results for Mask DINO and Mask2Former with 100 queries using a ResNet-50 backbone on ADE20K \texttt{val}. 
    We found the performance variance on this dataset is high and run three times to report both the mean and highest results for both models}.
    }
    \label{tab:sematic_ade}
\end{minipage}\hspace{10mm}
\begin{minipage}[t]{0.7\textwidth}
\makeatletter\def\@captype{table}
\centering
\large
\begin{tabular}{c|c|cccc}
    \toprule
        Model  & Iterations &\begin{tabular}[c]{@{}c@{}}mIoU\\(mean)\end{tabular} &\begin{tabular}[c]{@{}c@{}}mIoU\\(high)\end{tabular}&\begin{tabular}[c]{@{}c@{}}mIoU\\(reported)\end{tabular}  \\
        \midrule
        Mask2Former~\cite{cheng2021mask2former} &90k&$78.7$ & $79.0$&$79.4$   \\
          Mask DINO (ours)   & 90k &\textbf{79.8}\fontsize{7.0pt}{\baselineskip}\selectfont{(+1.1)} &\textbf{80.0}\fontsize{7.0pt}{\baselineskip}\selectfont{(+1.0)}&\textbf{80.0}\fontsize{7.0pt}{\baselineskip}\selectfont{(+0.6)}  \\
    \bottomrule
    \end{tabular}
    \centering
    \captionsetup{font={large}}
    \caption{Results for Mask DINO and Mask2Former with 100 queries using a ResNet-50 backbone on Cityscapes \texttt{val}. 
    We found the performance variance on this dataset is high and run three times to report both the mean and highest results for both models.
    }
    \label{tab:sematic_c}
\end{minipage}
\end{adjustbox}
\end{table*}

\begin{table*}[t]
    \centering
    \small
        \setlength\tabcolsep{2pt}
        \footnotesize
            \renewcommand{\arraystretch}{1.3}
    \begin{adjustbox}{width=0.9\textwidth,center}
    \begin{tabular}{c|c|c|c|c|cc}
        \toprule
        \multirow{2}{*}{Method} & \multirow{2}{*}{Params} & \multirow{2}{*}{Backbone} & \multirow{2}{*}{\makecell{Backbone Pre-training \\ Dataset }} &  \multirow{2}{*}{\makecell{Detection Pre-training \\ Dataset }}   & \multicolumn{2}{c}{{ val}}  \\
        & & &  && \scriptsize w/o TTA& \scriptsize w/ TTA  \\
        \midrule
        \multicolumn{5}{c}{{\textbf{Instance segmentation on COCO}}}&\multicolumn{2}{c}{AP} \\
        \midrule
        Mask2Former~\cite{cheng2021mask2former} & $216$M & SwinL & IN-22K-14M & $-$    & $50.1$ & $-$  \\
        Soft Teacher~\cite{xu2021end} & $284$M & SwinL & IN-22K-14M & O365     &$51.9$ & $52.5$  \\
        SwinV2-G-HTC++~\cite{liu2021swinv2} & $3.0$B & SwinV2-G & IN-22K-ext-70M~\cite{liu2021swinv2} & O365    & $53.4$ & $53.7$ \\
        \hline
        MasK DINO(Ours) & $\textbf{223}${M} & SwinL & IN-22K-14M & $-$ &    $\textbf{52.6}$ & ${-}$  \\
        MasK DINO(Ours) & $\textbf{223}${M} & SwinL & IN-22K-14M & O365 &    $\textbf{54.5}$\fontsize{7.0pt}{\baselineskip}\selectfont{(+1.1)} & ${-}$  \\
        
        \midrule
        \multicolumn{5}{c}{{\textbf{Panoptic segmentation on COCO}}}&\multicolumn{2}{c}{PQ} \\
       
        Panoptic SegFormer~\cite{li2021panoptic}& $-$M & SwinL & IN-22K-14M & $-$     &$55.8$ & $-$  \\
         Mask2Former~\cite{cheng2021mask2former} & $216$M & SwinL & IN-22K-14M & $-$     & $57.8$ & $-$  \\
        \hline
        MasK DINO (ours) & ${223}${M} & SwinL & IN-22K-14M &  &    $\textbf{58.4}$\fontsize{7.0pt}{\baselineskip}\selectfont{(+0.6)} & ${-}$  \\
        MasK DINO (ours) & ${223}${M} & SwinL & IN-22K-14M & O365 &    $\textbf{59.4}$\fontsize{7.0pt}{\baselineskip}\selectfont{(+1.6)} & ${-}$  \\
        \midrule
        \multicolumn{5}{c}{{\textbf{Semantic segmentation on ADE20K}}}&\multicolumn{2}{c}{mIoU} \\
        \midrule
        Mask2Former~\cite{cheng2021mask2former} & $215$M & SwinL & IN-22K-14M  & $-$     & $56.1$ & $57.3$  \\
        SeMask-L MSFaPN-Mask2Former~\cite{jain2021semask} & $-$M & SwinL-FaPN & IN-22K-14M & $-$     &$-$ & $58.2$  \\
        SwinV2-G-UperNet~\cite{liu2021swinv2} & $3.0$B & SwinV2-G & IN-22K-ext-70M~\cite{liu2021swinv2} & $-$ &   $59.3$ & $59.9$\\
        \hline
        MasK DINO (ours) & $\textbf{223}$\textbf{M} & SwinL & IN-22K-14M & $-$ &    $\textbf{56.6}$ & ${-}$  \\
        MasK DINO (ours) & $\textbf{223}$\textbf{M} & SwinL & IN-22K-14M & O365 &    $\textbf{59.5}$ & $\textbf{60.8}$\fontsize{7.0pt}{\baselineskip}\selectfont{(+0.9)}  \\
        \bottomrule
    \end{tabular}
    \end{adjustbox}
    \caption{Comparison of the SOTA models on three segmentation tasks. Mask DINO outperforms all existing models. "TTA" means test-time-augmentation.
    ``O365''  denotes the  Objects365~\cite{shao2019objects365} dataset. 
    }
    \label{tab:sota}
    \vspace{-.3cm}
\end{table*}

\section{Experiments}
We conduct extensive experiments and compare with several specialized models for four popular tasks including object detection, instance, panoptic, and semantic segmentation on COCO~\cite{lin2015microsoft}, ADE20K~\cite{zhou2017scene}, and Cityscapes~\cite{cordts2016cityscapes}. For all experiments, we use batch size 16 and A100 GPUs with 40GB memory. We use a ResNet-50~\cite{he2015deep} and a SwinL~\cite{liu2021swin} backbone for our main results and SOTA model.  Under ResNet-50, we use $4$ A100 GPUs for all tasks without extra data.
The implementation details are in Appendix \ref{sec:vis}.
\subsection{Main Results}\label{sec:main}
\noindent\textbf{Instance segmentation and object detection.} In Table ~\ref{tab:instance}, we compare Mask DINO with other instance segmentation and object detection models. Mask DINO outperforms both the specialized models such as Mask2Former~\cite{cheng2021mask2former} and DINO~\cite{zhang2022dino} and hybrid models such as HTC~\cite{chen2019hybrid} under the same setting. Especially, the instance segmentation results surpass the strong baseline Mask2Former by a large margin (+$\textbf{2.7}$ AP and +$\textbf{2.6}$ AP)  on the 12-epoch and 50-epoch settings. Moreover, Mask DINO significantly improves the convergence speed, outperforming Mask2Former with less than half training epochs (${44.2}$ AP in 24 epochs). 
In addition, after using mask-enhanced box initialization, our detection performance has been significantly improved (+1.2 AP), which even outperforms DINO by \textbf{0.8} AP. These results indicate that task unification is beneficial. Without bells and whistles, we achieve the best detection and instance segmentation performance among DETR-like model with a SwinL backbone without extra data.
\begin{table*}[t]
\begin{adjustbox}{width=1.0\textwidth,center}
\begin{minipage}[t]{0.55\textwidth}
\makeatletter\def\@captype{table}
\centering
\begin{tabular}{c|ccc}
    \toprule
    test layer\# & Mask DINO & Mask2Former  \\
        \midrule
        layer 0 & $\textbf{39.6}$\fontsize{7.0pt}{\baselineskip}\selectfont{(+38.5)} & $1.1$     \\
        \midrule
        layer 3&$\textbf{44.0}$&$42.3$\\
     layer 6     &$\textbf{45.9}$    & $43.3$  \\
     layer 9  &  $\textbf{46.0}$    & ${43.7}$  \\
    
    \bottomrule
\end{tabular}
\caption{Effectiveness of our query selection for mask initialization. We evaluate the instance segmentation performance from different decoder layers in the same model after training for 50 epochs. 
}
\label{tab:dec_test}
\end{minipage}\hspace{3mm}
\begin{minipage}[t]{0.38\textwidth}
\makeatletter\def\@captype{table}
\centering
\vspace{-41pt}
\begin{tabular}{c|cc|cccc}
    \toprule
    \multirow{2}{*}{layer\#}&\multicolumn{2}{c|}{w/o ME} & \multicolumn{2}{c}{w ME} \\
          &Box&  Mask &Box&  Mask \\
        \midrule
    layer 0    &   39.6   & \underline{25.6} &39.8&\textbf{41.2}\fontsize{7.0pt}{\baselineskip}\selectfont{(+15.6)} \\
     layer 9& 46.0  & \underline{50.5} & 46.3&\textbf{51.7}\fontsize{7.0pt}{\baselineskip}\selectfont{(+1.2)}   \\
    \bottomrule
\end{tabular}
\caption{Comparsion of our model with and without Mask-enhanced anchor box initialization (ME). ME enhances anchor box initialization and improves final detection performance. Trained for 50 epochs.
}
\label{tab:maskenhance}
\end{minipage}\hspace{6mm}
\begin{minipage}[t]{0.42\textwidth}
\makeatletter\def\@captype{table}
\centering
\vspace{-40.5pt}
\begin{tabular}{c|ccc}
    \toprule
   Feature scale & box AP & {mask AP}  \\
        \midrule
    single scale(1/8)     &  $45.8$    & $45.1$  \\
     3 scales  & $50.5$ & $45.8$   \\
    4 scales & $50.5$ & $\textbf{46.0}$     \\
    \bottomrule
\end{tabular}
\caption{Comparison of multi-scale features for Transformer decoder under the 50-epoch setting. Both detection and segmentation benefit from more feature scales.}
\label{tab:feature_scale}
\end{minipage}\hspace{3mm}

\end{adjustbox}
\vspace{-.3cm}
\end{table*}

\begin{table*}[t]
\begin{adjustbox}{width=1.0\textwidth,center}
\begin{minipage}[t]{0.46\textwidth}
\makeatletter\def\@captype{table}
\centering
\begin{tabular}{cc|cc}
    \toprule
    \multicolumn{2}{c|}{Tasks} & \multirow{2}{*}{Box AP} & \multirow{2}{*}{Mask AP} \\
          Box&  Mask & & \\
        \midrule
    \checkmark     &      & $50.1$ &$-$ \\
     & \checkmark  & $-$ & $43.3$   \\
    \checkmark     & \checkmark & $\textbf{50.5}$\fontsize{7.0pt}{\baselineskip}\selectfont{(+0.4)} & $\textbf{46.0}$\fontsize{7.0pt}{\baselineskip}\selectfont{(+2.7)}     \\
    \bottomrule
\end{tabular}
\caption{Task comparison under the 50-epoch setting. We train the same Mask DINO with different tasks and validate that box and mask can achieve mutual cooperation.}
\label{tab:taskhelp}
\end{minipage}\hspace{6mm}
\begin{minipage}[t]{0.48\textwidth}
\makeatletter\def\@captype{table}
\centering
\begin{tabular}{c|ccc}
    \toprule
   Decoder layer\# & Box AP & {Mask AP}  \\
        \midrule
        3&$43.1$&$40.7$\\
     6     &$44.3$    & $41.1$  \\
     9  &  $44.5$    & $\textbf{41.4}$  \\
    12 & $44.8$ & $41.1$     \\
    
    \bottomrule
\end{tabular}
\caption{Decoder layer number comparison under the 12-epoch setting. Mask DINO benefits from more decoders, while DINO's performance will decrease with 9 decoders.}
\label{tab:dec_layer}
\end{minipage}\hspace{6mm}

\begin{minipage}[t]{0.48\textwidth}
\makeatletter\def\@captype{table}
\centering
\vspace{-45pt}
\begin{tabular}{cc|cc}
    \toprule
    \multicolumn{2}{c}{Matching} & \multirow{2}{*}{Box AP} & \multirow{2}{*}{Mask AP} \\
          Box&  Mask & & \\
        \midrule
    \checkmark     &      & $44.4$ & $40.5$ \\
     & \checkmark  & $40.2$ &$38.4$     \\
    \checkmark & \checkmark     & $\textbf{44.5}$ & $\textbf{41.4}$      \\
    \bottomrule
\end{tabular}
\caption{Matching method comparison under the 12-epoch setting. We train both tasks together but use different matching methods to verify the effectiveness of hybrid matching.}
\label{tab:match}
\end{minipage}\hspace{6mm}
\end{adjustbox}
\vspace{-.4cm}
\end{table*}


\begin{table*}[!t]
\begin{adjustbox}{width=1.0\textwidth,center}
\begin{minipage}[t]{0.8\textwidth}
\makeatletter\def\@captype{table}
\centering
\centering
\vspace{-37.005pt}
\begin{tabular}{c|c|ccccc}
    \toprule
    &Epochs& PQ & PQ$^{thing}$& PQ$^{stf}$&Box AP$^{Th}_{pan}$&Mask AP$^{Th}_{pan}$ \\
        \midrule
        w/o decouple&12&$47.9$&$54.0$&$38.8$&$42.8$&$39.6$\\
     w/ decouple    &12 &$\textbf{49.0}$\fontsize{7.0pt}{\baselineskip}\selectfont{(+1.1)}    & $\textbf{54.8}$& $\textbf{40.2}$& $\textbf{43.2}$& $\textbf{40.4}$  \\
     \midrule
     w/o decouple  &50&  $52.7$  & $58.8$&$43.5$&$48.7$&$44.1$  \\
    w/ decouple &50 &\textbf{53.0}\fontsize{7.0pt}{\baselineskip}\selectfont{(+0.3)} &\textbf{59.1} & $\textbf{43.9}$ & $\textbf{48.8}$ & \textbf{44.3}     \\
    \bottomrule
\end{tabular}
\caption{Effectiveness of decoupled box prediction for panoptic segmentation under the 12-epoch and 50-epoch settings.}
\label{tab:decouple}
\end{minipage}

\begin{minipage}[t]{0.65\textwidth}
\makeatletter\def\@captype{table}
\centering
    \begin{tabular}{l|ccc}
        \toprule
          & Box AP & Mask AP   \\
        \midrule
         Mask DINO (ours) & \textbf{45.7}&\textbf{41.4} \\
         \midrule
        $-$ Mask-enhanced anchor box initialization  & 44.5\fontsize{7.0pt}{\baselineskip}\selectfont{(-1.2)}&41.4    \\
         \quad$-$ Unified query selection for masks  & 43.6\fontsize{7.0pt}{\baselineskip}\selectfont{(-2.1)} &40.3\fontsize{7.0pt}{\baselineskip}\selectfont{(-1.1)}   \\
        $-$ Unified denoising for masks & 44.4\fontsize{7.0pt}{\baselineskip}\selectfont{(-1.3)}&40.3 \fontsize{7.0pt}{\baselineskip}\selectfont{(-1.1)}    \\
        $-$ Hybrid matching & 44.9\fontsize{7.0pt}{\baselineskip}\selectfont{(-0.8)} & 40.5 \fontsize{7.0pt}{\baselineskip}\selectfont{(-0.9)}   \\
        \midrule
        $-$ remove all the above & 41.7 \fontsize{7.0pt}{\baselineskip}\selectfont{(-4.0)} & 38.5 \fontsize{7.0pt}{\baselineskip}\selectfont{(-2.7)}   \\
        \bottomrule
    \end{tabular}
    \caption{Effectiveness of the proposed components under the 12-epoch setting. 
    }
    \label{tab:ablation}
    \end{minipage}
    \end{adjustbox}
    \vspace{-.5cm}
\end{table*}

\noindent\textbf{Panoptic segmentation.} We compare Mask DINO with other models in Table~\ref{tab:panoptic}. Mask DINO outperforms all previous best models on both the $12$-epoch and $50$-epoch settings by $\textbf{1.0}$ PQ and $\textbf{1.1}$ PQ, respectively. This indicates Mask DINO has the advantages of both faster convergence and superior performance. One interesting observation is that we outperform Mask2Former~\cite{cheng2021mask2former} in terms of both $PQ^{Th}$ and $PQ^{St}$. However, instead of using dense and hard-constrained masked attention, we predict boxes and then use them in deformable attention to extract query features. Therefore, our box-oriented deformable attention also works well with "stuff" categories,  which makes our unified model simple and efficient. 
In addition, we improve the mask AP$_{pan}^{Th}$ by $\textbf{2.6}$ to $44.3$ AP, which is $0.6$ higher than the specialized instance segmentation model Mask2Fomer ($43.7$ AP).
\\\textbf{Semantic segmentation.} In Table ~\ref{tab:sematic_ade} and \ref{tab:sematic_c}, we show the performance of semantic segmentation with a ResNet-50 backbone. We use $100$ queries for these small datasets. We outperform Mask2Former on both ADE20K and Cityscapes by ${1.6}$ and ${0.6}$ mIoU on the reported performance.

\subsection{Comparison with SOTA Models}
In Table ~\ref{tab:sota}, we compare Mask DINO with SOTA models on three image segmentation tasks to show its scalability. We use the SwinL~\cite{liu2021swin} backbone and pre-train DINO on the Objects365~\cite{shao2019objects365} detection dataset. Even without using extra data, we outperform Mask2Former on all three tasks, especially on instance segmentation (\textbf{+2.5 AP}). As Mask DINO is an extension of DINO, the pre-trained DINO model can be used to fine-tune Mask DINO for segmentation tasks. After fine-tuning Mask DINO on the corresponding tasks, we achieve the best results on instance ($\textbf{54.5}$ AP), panoptic ($\textbf{59.4}$  PQ), and semantic ($\textbf{60.8}$ mIoU) segmentation among model under one billion parameters. Compared to SwinV2-G~\cite{liu2021swinv2}, we significantly reduce the model size to 1/15 and backbone pre-training dataset to 1/5. Our detection pre-training also significantly helps all segmentation tasks including panoptic and semantic with "stuff" categories. However, previous specialized segmentation models such as Mask2Former can not use detection datasets and adding a detection head to it results in poor performance as shown in Table \ref{tab:mask2former_head}, which severely limits the data scalability. By unifying four tasks in one model, we only need to pre-train one model on a large-scale dataset and finetune on all tasks for 10 to 20 epochs (Mask2Former needs 100 epochs), which is more computationally efficient and simpler in model design.

\subsection{Ablation Studies}
We conduct ablation studies using a ResNet-50 backbone to analyze Mask DINO on COCO \texttt{val2017}. Unless otherwise stated, our experiments are based on object detection and instance segmentation without \emph{Mask-enhanced anchor box initialization}.
\\\textbf{Query selection.} Table ~\ref{tab:dec_test} shows the results of our query selection for instance segmentation, where we additionally provide the performance of different decoder layers in one single model. Mask2Former also predicts the masks of learnable queries as initial region proposals. However, their performance lags behind Mask DINO by a large margin ($\textbf{-38.5} AP$). With our effective query selection scheme, the mask performance achieves $\textbf{39.6}$ AP without using the decoder. In addition, our mask performance at layer six is already comparable to the final results with 9 layers. In Table \ref{tab:maskenhance}, we show that in query selection the predicted box is inferior to mask, which indicates segmentation is easier to learn in the initial stage. Therefore, our proposed mask-enhanced box initialization enhances boxes with masks in query selection to provide better anchor boxes (+15.6 AP) for the decoder, which results in \textbf{+1.2} AP improvement in the final detection performance. 
\\\textbf{Feature scales.}
Mask2Former~\cite{cheng2021mask2former} shows that concatenating multi-scale features as input to Transformer decoder layers does not improve the segmentation performance. However, in Table \ref{tab:feature_scale}, Mask DINO shows that using more feature scales in the decoder consistently improves the performance.
\\\textbf{Object detection and segmentation help each other.} To validate task cooperation in Mask DINO, we use the same model but train different tasks and report the 12 epoch and 50 epoch results. As shown in Table ~\ref{tab:taskhelp}, only training one task will lead to a performance drop. Although only training object detection results in faster convergence in the early stage for box prediction, the final  performance is still inferior to training both tasks together.
\\\textbf{Decoder layer number.}
In DINO, increasing the decoder layer number to nine will decrease the performance of box. In Table \ref{tab:dec_layer}, the result indicates that increasing the number of decoder layers will contribute to both detection and segmentation in Mask DINO. We hypothesize that the multi-task training become more complex and require more decoders to learn the needed mapping function.
\\\textbf{Matching.}
In Table~\ref{tab:match}, we show that only using boxes or masks to perform bipartite matching is not optimal in Mask DINO. A unified matching objective makes the optimization more consistent.
\\\textbf{Decoupled box prediction.}
In Table~\ref{tab:decouple}, we show the effectiveness of our decoupled box prediction for panoptic segmentation. This decoupled design of "thing" and "stuff" accelerates training in the early stage (12-epoch setting) and improves the final performance (50-epoch setting).
\\\textbf{Effectiveness of the algorithm components.} In Table \ref{tab:ablation}, we remove each algorithm component at a time and show that each component contributes to the final performance. 
In addition, after removing all the proposed components, both detection and segmentation performance drop by a large margin. This result indicates that if we trivially add detection and segmentation tasks in one DETR-based model, the features are not aligned for detection and segmentation tasks to achieve mutual cooperation.

We also present visualization analysis in Appendix \ref{sec:vis}.

\section{Conclusion}
In this paper, we have presented Mask DINO as a unified Transformer-based framework for both object detection and image segmentation. Conceptually, Mask DINO is a natural extension of DINO from detection to segmentation with minimal modifications on some key components.
Mask DINO outperforms previous specialized models and achieves the best results on all three segmentation tasks (instance, panoptic, and semantic) among models under one billion parameters. {Moreover, Mask DINO shows that detection and segmentation can help each other in query-based models.} In particular, Mask DINO enables semantic and panoptic segmentation to benefit from a better visual representation pre-trained on a large-scale detection dataset. We hope Mask DINO can provide insights for enabling task cooperation and data cooperation towards designing a universal model for more vision tasks.
\\\textbf{Limitations: }Different segmentation tasks fail to achieve mutual assistance in Mask DINO in COCO panoptic segmentation. 
For example, in COCO panoptic segmentation, the mask AP still lags behind the model only trained with instances. 
In addition, under the large-scale setting, we have not achieved a new SOTA detection performance as the segmentation head requires additional GPU memory. To accommodate this memory limitation, for the large-scale setting, we have to use smaller image size and less number of queries compared with DINO, which impacts the final performance of object detection. In the future, we will further optimize the implementation to develop a more universal and efficient model to promote task cooperation.

{\small
\bibliographystyle{ieee_fullname}
\bibliography{egbib}

\begin{thebibliography}{10}\itemsep=-1pt

\bibitem{carion2020end}
Nicolas Carion, Francisco Massa, Gabriel Synnaeve, Nicolas Usunier, Alexander
  Kirillov, and Sergey Zagoruyko.
\newblock {End-to-end object detection with transformers}.
\newblock In {\em European conference on computer vision}, pages 213--229.
  Springer, 2020.

\bibitem{chen2019hybrid}
Kai Chen, Jiangmiao Pang, Jiaqi Wang, Yu Xiong, Xiaoxiao Li, Shuyang Sun,
  Wansen Feng, Ziwei Liu, Jianping Shi, Wanli Ouyang, et~al.
\newblock {Hybrid task cascade for instance segmentation}.
\newblock In {\em Proceedings of the IEEE/CVF Conference on Computer Vision and
  Pattern Recognition}, pages 4974--4983, 2019.

\bibitem{cheng2021mask2former}
Bowen Cheng, Ishan Misra, Alexander~G. Schwing, Alexander Kirillov, and Rohit
  Girdhar.
\newblock {Masked-attention Mask Transformer for Universal Image Segmentation}.
\newblock 2022.

\bibitem{cheng2021pointly}
Bowen Cheng, Omkar Parkhi, and Alexander Kirillov.
\newblock {Pointly-supervised instance segmentation}.
\newblock {\em arXiv preprint arXiv:2104.06404}, 2021.

\bibitem{cheng2021maskformer}
Bowen Cheng, Alexander~G. Schwing, and Alexander Kirillov.
\newblock {Per-Pixel Classification is Not All You Need for Semantic
  Segmentation}.
\newblock 2021.

\bibitem{cordts2016cityscapes}
Marius Cordts, Mohamed Omran, Sebastian Ramos, Timo Rehfeld, Markus Enzweiler,
  Rodrigo Benenson, Uwe Franke, Stefan Roth, and Bernt Schiele.
\newblock {The cityscapes dataset for semantic urban scene understanding}.
\newblock In {\em Proceedings of the IEEE conference on computer vision and
  pattern recognition}, pages 3213--3223, 2016.

\bibitem{du2021simple}
Xianzhi Du, Barret Zoph, Wei-Chih Hung, and Tsung-Yi Lin.
\newblock {Simple training strategies and model scaling for object detection}.
\newblock {\em arXiv preprint arXiv:2107.00057}, 2021.

\bibitem{everingham2015pascal}
Mark Everingham, SM Eslami, Luc Van~Gool, Christopher~KI Williams, John Winn,
  and Andrew Zisserman.
\newblock {The pascal visual object classes challenge: A retrospective}.
\newblock {\em International journal of computer vision}, 111(1):98--136, 2015.

\bibitem{fang2021instances}
Yuxin Fang, Shusheng Yang, Xinggang Wang, Yu Li, Chen Fang, Ying Shan, Bin
  Feng, and Wenyu Liu.
\newblock {Instances as queries}.
\newblock In {\em Proceedings of the IEEE/CVF International Conference on
  Computer Vision}, pages 6910--6919, 2021.

\bibitem{ghiasi2021simple}
Golnaz Ghiasi, Yin Cui, Aravind Srinivas, Rui Qian, Tsung-Yi Lin, Ekin~D Cubuk,
  Quoc~V Le, and Barret Zoph.
\newblock {Simple copy-paste is a strong data augmentation method for instance
  segmentation}.
\newblock In {\em Proceedings of the IEEE/CVF Conference on Computer Vision and
  Pattern Recognition}, pages 2918--2928, 2021.

\bibitem{he2017mask}
Kaiming He, Georgia Gkioxari, Piotr Doll{\'a}r, and Ross Girshick.
\newblock {Mask r-cnn}.
\newblock In {\em Proceedings of the IEEE international conference on computer
  vision}, pages 2961--2969, 2017.

\bibitem{he2015deep}
Kaiming {He}, Xiangyu {Zhang}, Shaoqing {Ren}, and Jian {Sun}.
\newblock Deep residual learning for image recognition.
\newblock In {\em 2016 IEEE Conference on Computer Vision and Pattern
  Recognition (CVPR)}, pages 770--778, 2016.

\bibitem{jain2022oneformer}
Jitesh Jain, Jiachen Li, MangTik Chiu, Ali Hassani, Nikita Orlov, and Humphrey
  Shi.
\newblock {OneFormer: One Transformer to Rule Universal Image Segmentation}.
\newblock {\em arXiv preprint arXiv:2211.06220}, 2022.

\bibitem{jain2021semask}
Jitesh Jain, Anukriti Singh, Nikita Orlov, Zilong Huang, Jiachen Li, Steven
  Walton, and Humphrey Shi.
\newblock {SeMask: Semantically Masked Transformers for Semantic Segmentation}.
\newblock {\em arXiv preprint arXiv:2112.12782}, 2021.

\bibitem{kirillov2019pafpn}
Alexander Kirillov, Ross Girshick, Kaiming He, and Piotr Doll{\'a}r.
\newblock {Panoptic feature pyramid networks}.
\newblock In {\em Proceedings of the IEEE/CVF Conference on Computer Vision and
  Pattern Recognition}, pages 6399--6408, 2019.

\bibitem{kirillov2019panoptic}
Alexander Kirillov, Kaiming He, Ross Girshick, Carsten Rother, and Piotr
  Doll{\'a}r.
\newblock {Panoptic segmentation}.
\newblock In {\em Proceedings of the IEEE/CVF Conference on Computer Vision and
  Pattern Recognition}, pages 9404--9413, 2019.

\bibitem{kirillov2020pointrend}
Alexander Kirillov, Yuxin Wu, Kaiming He, and Ross Girshick.
\newblock {Pointrend: Image segmentation as rendering}.
\newblock In {\em Proceedings of the IEEE/CVF conference on computer vision and
  pattern recognition}, pages 9799--9808, 2020.

\bibitem{li2022dn}
Feng Li, Hao Zhang, Shilong Liu, Jian Guo, Lionel~M Ni, and Lei Zhang.
\newblock {DN-DETR: Accelerate DETR Training by Introducing Query DeNoising}.
\newblock {\em arXiv preprint arXiv:2203.01305}, 2022.

\bibitem{li2021panoptic}
Zhiqi Li, Wenhai Wang, Enze Xie, Zhiding Yu, Anima Anandkumar, Jose~M Alvarez,
  Tong Lu, and Ping Luo.
\newblock {Panoptic SegFormer}.
\newblock {\em arXiv preprint arXiv:2109.03814}, 2021.

\bibitem{lin2018focal}
Tsung-Yi {Lin}, Priya {Goyal}, Ross {Girshick}, Kaiming {He}, and Piotr
  {Dollar}.
\newblock Focal loss for dense object detection.
\newblock {\em IEEE Transactions on Pattern Analysis and Machine Intelligence},
  42(2):318--327, 2020.

\bibitem{lin2015microsoft}
Tsung-Yi Lin, Michael Maire, Serge Belongie, James Hays, Pietro Perona, Deva
  Ramanan, Piotr Doll{\'a}r, and C~Lawrence Zitnick.
\newblock {Microsoft COCO: Common objects in context}.
\newblock In {\em European conference on computer vision}, pages 740--755.
  Springer, 2014.

\bibitem{liu2022dab}
Shilong Liu, Feng Li, Hao Zhang, Xiao Yang, Xianbiao Qi, Hang Su, Jun Zhu, and
  Lei Zhang.
\newblock {DAB-DETR: Dynamic Anchor Boxes are Better Queries for DETR}.
\newblock {\em arXiv preprint arXiv:2201.12329}, 2022.

\bibitem{liu2021swinv2}
Ze Liu, Han Hu, Yutong Lin, Zhuliang Yao, Zhenda Xie, Yixuan Wei, Jia Ning, Yue
  Cao, Zheng Zhang, Li Dong, et~al.
\newblock {Swin Transformer V2: Scaling Up Capacity and Resolution}.
\newblock {\em arXiv preprint arXiv:2111.09883}, 2021.

\bibitem{liu2021swin}
Ze Liu, Yutong Lin, Yue Cao, Han Hu, Yixuan Wei, Zheng Zhang, Stephen Lin, and
  Baining Guo.
\newblock {Swin transformer: Hierarchical vision transformer using shifted
  windows}.
\newblock In {\em Proceedings of the IEEE/CVF International Conference on
  Computer Vision}, pages 10012--10022, 2021.

\bibitem{long2015fully}
Jonathan Long, Evan Shelhamer, and Trevor Darrell.
\newblock {Fully convolutional networks for semantic segmentation}.
\newblock In {\em Proceedings of the IEEE conference on computer vision and
  pattern recognition}, pages 3431--3440, 2015.

\bibitem{meng2021conditional}
Depu Meng, Xiaokang Chen, Zejia Fan, Gang Zeng, Houqiang Li, Yuhui Yuan, Lei
  Sun, and Jingdong Wang.
\newblock {Conditional DETR for Fast Training Convergence}.
\newblock {\em arXiv preprint arXiv:2108.06152}, 2021.

\bibitem{qin2022pyramid}
Zipeng Qin, Jianbo Liu, Xiaolin Zhang, Maoqing Tian, Aojun Zhou, Shuai Yi, and
  Hongsheng Li.
\newblock {Pyramid Fusion Transformer for Semantic Segmentation}.
\newblock {\em arXiv preprint arXiv:2201.04019}, 2022.

\bibitem{ren2015faster}
Shaoqing Ren, Kaiming He, Ross Girshick, and Jian Sun.
\newblock {Faster r-cnn: Towards real-time object detection with region
  proposal networks}.
\newblock {\em Advances in neural information processing systems}, 28, 2015.

\bibitem{rezatofighi2019generalized}
Hamid Rezatofighi, Nathan Tsoi, JunYoung Gwak, Amir Sadeghian, Ian Reid, and
  Silvio Savarese.
\newblock {Generalized intersection over union: A metric and a loss for
  bounding box regression}.
\newblock In {\em Proceedings of the IEEE/CVF Conference on Computer Vision and
  Pattern Recognition}, pages 658--666, 2019.

\bibitem{ronneberger2015u}
Olaf Ronneberger, Philipp Fischer, and Thomas Brox.
\newblock {U-net: Convolutional networks for biomedical image segmentation}.
\newblock In {\em International Conference on Medical image computing and
  computer-assisted intervention}, pages 234--241. Springer, 2015.

\bibitem{shao2019objects365}
Shuai Shao, Zeming Li, Tianyuan Zhang, Chao Peng, Gang Yu, Xiangyu Zhang, Jing
  Li, and Jian Sun.
\newblock {Objects365: A large-scale, high-quality dataset for object
  detection}.
\newblock In {\em Proceedings of the IEEE/CVF international conference on
  computer vision}, pages 8430--8439, 2019.

\bibitem{vaswani2017attention}
Ashish Vaswani, Noam Shazeer, Niki Parmar, Jakob Uszkoreit, Llion Jones,
  Aidan~N Gomez, {\L}ukasz Kaiser, and Illia Polosukhin.
\newblock {Attention is all you need}.
\newblock In {\em Advances in neural information processing systems}, pages
  5998--6008, 2017.

\bibitem{wang2021max}
Huiyu Wang, Yukun Zhu, Hartwig Adam, Alan Yuille, and Liang-Chieh Chen.
\newblock {Max-deeplab: End-to-end panoptic segmentation with mask
  transformers}.
\newblock In {\em Proceedings of the IEEE/CVF Conference on Computer Vision and
  Pattern Recognition}, pages 5463--5474, 2021.

\bibitem{wang2021anchor}
Yingming {Wang}, Xiangyu {Zhang}, Tong {Yang}, and Jian {Sun}.
\newblock Anchor detr: Query design for transformer-based detector.
\newblock {\em arXiv preprint arXiv:2109.07107}, 2021.

\bibitem{xiong2019upsnet}
Yuwen Xiong, Renjie Liao, Hengshuang Zhao, Rui Hu, Min Bai, Ersin Yumer, and
  Raquel Urtasun.
\newblock {Upsnet: A unified panoptic segmentation network}.
\newblock In {\em Proceedings of the IEEE/CVF Conference on Computer Vision and
  Pattern Recognition}, pages 8818--8826, 2019.

\bibitem{xu2021end}
Mengde Xu, Zheng Zhang, Han Hu, Jianfeng Wang, Lijuan Wang, Fangyun Wei, Xiang
  Bai, and Zicheng Liu.
\newblock {End-to-end semi-supervised object detection with soft teacher}.
\newblock In {\em Proceedings of the IEEE/CVF International Conference on
  Computer Vision}, pages 3060--3069, 2021.

\bibitem{zhang2022dino}
Hao Zhang, Feng Li, Shilong Liu, Lei Zhang, Hang Su, Jun Zhu, Lionel~M Ni, and
  Heung-Yeung Shum.
\newblock {DINO: DETR with Improved DeNoising Anchor Boxes for End-to-End
  Object Detection}.
\newblock {\em arXiv preprint arXiv:2203.03605}, 2022.

\bibitem{zhang2021k}
Wenwei Zhang, Jiangmiao Pang, Kai Chen, and Chen~Change Loy.
\newblock {K-net: Towards unified image segmentation}.
\newblock {\em Advances in Neural Information Processing Systems}, 34, 2021.

\bibitem{zhou2017scene}
Bolei Zhou, Hang Zhao, Xavier Puig, Sanja Fidler, Adela Barriuso, and Antonio
  Torralba.
\newblock {Scene parsing through ade20k dataset}.
\newblock In {\em Proceedings of the IEEE conference on computer vision and
  pattern recognition}, pages 633--641, 2017.

\bibitem{zhu2020deformable}
Xizhou {Zhu}, Weijie {Su}, Lewei {Lu}, Bin {Li}, Xiaogang {Wang}, and Jifeng
  {Dai}.
\newblock Deformable detr: Deformable transformers for end-to-end object
  detection.
\newblock In {\em ICLR 2021: The Ninth International Conference on Learning
  Representations}, 2021.

\end{thebibliography}
}

\newpage
\appendix
\newpage
\clearpage

\setcounter{page}{1}
\section{Visualization analysis}\label{sec:vis}
\begin{figure*}[htbp]
\centering
\begin{minipage}[h]{0.12\textwidth}
\includegraphics[width=0.75in]{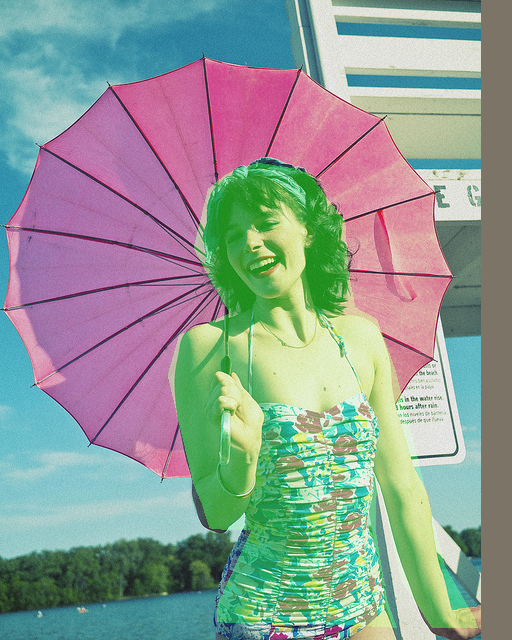}
\subcaption{}
\end{minipage}%
\begin{minipage}[h]{0.12\textwidth}
\includegraphics[width=0.75in]{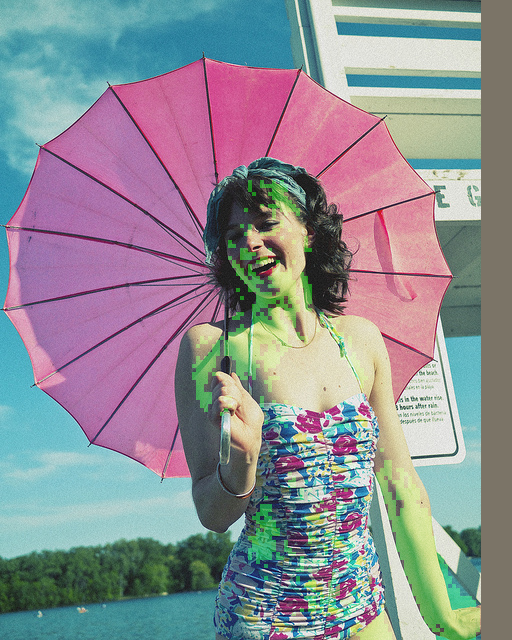}
\subcaption{}
\end{minipage}%
\begin{minipage}[h]{0.12\linewidth}
\includegraphics[width=0.75in]{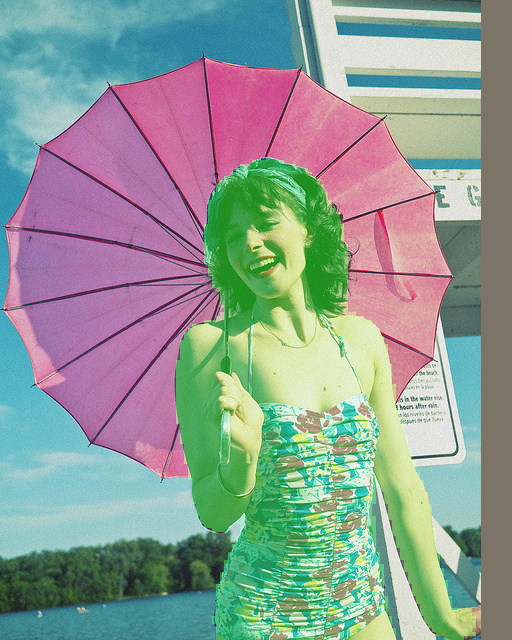}
\subcaption{}
\end{minipage}
\begin{minipage}[h]{0.12\linewidth}
\includegraphics[width=0.75in]{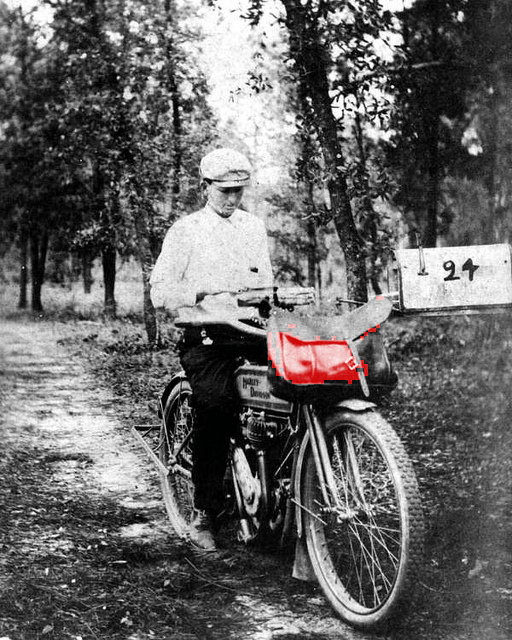}
\subcaption{}
\end{minipage}
\begin{minipage}[h]{0.12\linewidth}
\includegraphics[width=0.75in]{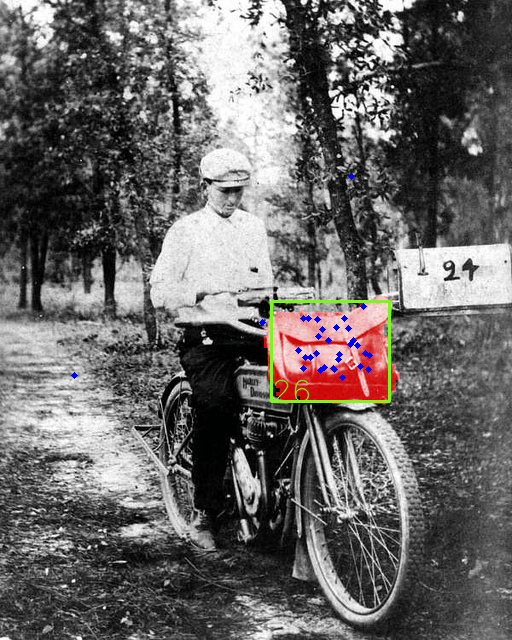}
\subcaption{}
\end{minipage}
\begin{minipage}[h]{0.15\linewidth}
\includegraphics[width=1.05in]{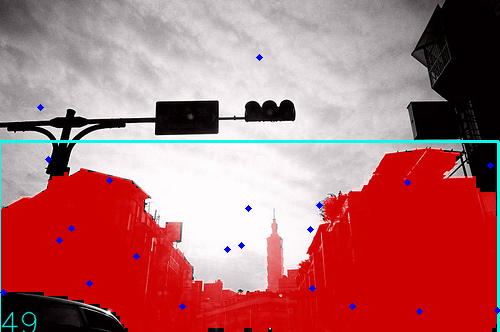}
\subcaption{}
\end{minipage}
\begin{minipage}[h]{0.15\linewidth}
\includegraphics[width=1.05in]{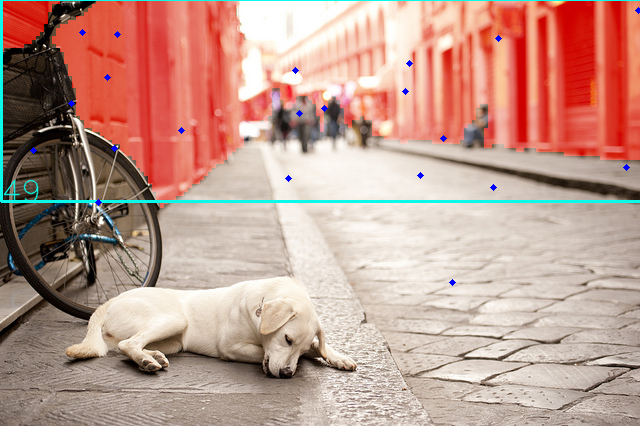}
\subcaption{}
\end{minipage}

\caption{(a) The green transparent region is the ground truth mask for the girl. (b)(c) The predicted masks of the $0$-th decoder layer in Mask2Former and Mask DINO, respectively. Note that we attain the predicted masks by first choosing the query which is finally assigned to the ground truth mask in the last decoder layer. Then we visualize the predicted mask of this query by performing dot production with the pixel embedding map. (d)(e) The outputs of the $1$-st layer in Mask2Former and Mask DINO. The red masks are predicted masks and the green box is the predicted box by Mask DINO. The blue points are sampled points by deformable attention. Since the $0$-th layer of Mask2Former usually outputs unfavorable masks, we avoid using its $0$-th layer here. (f)(g) show that Mask DINO can predict correct sampled points, boxes, and masks for background stuffs.}
\label{fig:analyze_2st}
\end{figure*}

There has been a trend to unify detection and segmentation tasks using convolution-based models, which not only simplifies model design but also promotes mutual cooperation between detection and segmentation. There are mainly three motivations for us to propose Mask DINO. \textbf{First}, DINO~\cite{zhang2022dino} has achieved SOTA results on object detection. Previous works such as Mask RCNN~\cite{he2017mask}, HTC~\cite{chen2019hybrid}, and DETR~\cite{carion2020end} have shown that a detection model can be extended to do segmentation and help design better segmentation models. 
\textbf{Second}, detection is a relatively easier task than instance segmentation. As shown in Table   \ref{tab:instance} (and other previous studies), Box AP is usually $4+$ AP higher than mask AP. Therefore, box prediction can guide attention to focus on more meaningful regions and extract better features for mask prediction. \textbf{Third}, the new improvements in DINO and other DETR-like models~\cite{zhu2020deformable,li2022dn} such as query selection and deformable attention can also help segmentation tasks. For example, Mask2Former adopts learnable decoder queries, which cannot take advantage of the position information in the selected top $K$ features from the encoder to guide mask predictions. Fig.   \ref{fig:analyze_2st}(a)(b)(c) show that the output of Mask2Former in the $0$-th decoder layer is far away from the GT mask while Mask DINO outputs much better masks as region proposals. Mask2Former also adopts specialized masked attention to guide the model to attend to regions of interest. However, masked attention is a hard constraint which ignores features outside a provided mask and may overlook important information for following decoder layers. In addition, deformable attention is also a better substitute for its high efficiency allowing attention to be applied to multi-scale features without too much computational overhead. Fig.     \ref{fig:analyze_2st}(d)(e) show a predicted mask of Mask2Former in its $1$-st decoder layer and the corresponding output of Mask DINO. The prediction of Mask2Former only covers less than half of the GT mask, which means that the attention can not see the whole instance in the next decoder layer. 
Moreover, a box can also guide deformable attention to a proper region for background stuff, as shown in Fig.   \ref{fig:analyze_2st}(f)(g).

\section{Implementation details}\label{sec:impl}
\textbf{The code is available in the supplementary materials}. We also provide some detailed descriptions of our implementation here.
\subsection{General settings}\label{sec:generalimpl}
\textbf{Dataset and metrics: }We evaluate Mask DINO on two challenging datasets: COCO 2017~\cite{lin2015microsoft} for object detection, instance segmentation, and panoptic segmentation; ADE20K~\cite{zhou2017scene} for semantic segmentation. They both have "thing" and "stuff" categories, therefore we follow the common practice to evaluate object detection and instance segmentation on the "thing" categories and evaluate panoptic and semantic segmentation on the union of the "thing" and "stuff" categories. Unless otherwise stated, all results are trained on the \texttt{train} split and evaluated on the \texttt{validation} split. For object detection and instance segmentation, the results are evaluated with the standard average precision (AP) and mask AP~\cite{lin2015microsoft} result. For panoptic segmentation, we evaluate the results with the panoptic quality (PQ) metric~\cite{kirillov2019panoptic}. We also report $AP_{pan}^{Th}$ (AP on the "thing" categories) and $AP_{pan}^{St}$ (AP on the "stuff" categories). For semantic segmentation, the results are evaluated with the mean Intersection-over-Union (mIoU) metric~\cite{everingham2015pascal}.
\begin{table*}[h]
    \centering
    \small
        \setlength\tabcolsep{2pt}
        \footnotesize
            \renewcommand{\arraystretch}{1.3}
    \begin{adjustbox}{width=0.8\textwidth,center}
    \begin{tabular}{c|c|c|c|c|cc}
        \toprule
        \multirow{2}{*}{Method} & \multirow{2}{*}{Params} & \multirow{2}{*}{Backbone} & \multirow{2}{*}{\makecell{Backbone Pre-training \\ Dataset }} &  \multirow{2}{*}{\makecell{Detection Pre-training \\ Dataset }}   & \multicolumn{2}{c}{{ test}}  \\
        & & &  && \scriptsize w/o TTA& \scriptsize w/ TTA  \\
        \midrule
        \multicolumn{5}{c}{{\textbf{Instance segmentation on COCO}}}&\multicolumn{2}{c}{AP} \\
        \midrule
        Mask2Former~\cite{cheng2021mask2former} & $216$M & SwinL & IN-22K-14M & $-$    & $50.5$ & $-$  \\
        Soft Teacher~\cite{xu2021end} & $284$M & SwinL & IN-22K-14M & O365     &- & $53.0$  \\
        SwinV2-G-HTC++~\cite{liu2021swinv2} & $3.0$B & SwinV2-G & IN-22K-ext-70M~\cite{liu2021swinv2} & O365    & - & $54.4$ \\
        \hline
        MasK DINO(Ours) & $\textbf{223}${M} & SwinL & IN-22K-14M & O365 &    $\textbf{54.7}$ & ${-}$  \\
        \midrule
        \multicolumn{5}{c}{{\textbf{Panoptic segmentation on COCO}}}&\multicolumn{2}{c}{PQ} \\
        \midrule
       
        Panoptic SegFormer~\cite{li2021panoptic}& $-$M & SwinL & IN-22K-14M & $-$     &$56.2$ & $-$  \\
         Mask2Former~\cite{cheng2021mask2former} & $216$M & SwinL & IN-22K-14M & $-$     & $58.3$ & $-$  \\

        \hline
        MasK DINO (ours) & ${223}${M} & SwinL & IN-22K-14M & O365 &    $\textbf{59.5}$& ${-}$  \\
        \bottomrule
    \end{tabular}
    \end{adjustbox}
    \caption{Comparison of SOTA models on COCO \texttt{test-dev}. Mask DINO outperforms all existing models. "TTA" means test-time-augmentation.
    ``O365''  denotes the  Objects365~\cite{shao2019objects365} dataset. 
    }
    \label{tab:testsota}
    \vspace{-.7cm}
\end{table*}
\\\textbf{Backbone: }We report results with two public backbones: ResNet-50~\cite{he2015deep} and SwinL~\cite{liu2021swin}. To achieve SOTA performance using a large model with the SwinL backbone, we use Objects365~\cite{shao2019objects365} to pre-train an object detection model and then fine-tune the model on the corresponding datasets for all tasks. Though we only pre-train for object detection, our model generalizes well to improve the performance of all segmentation tasks.
\\\textbf{Loss function: }As we train detection and segmentation tasks jointly, there are totally three kinds of losses, including classification loss $\mathcal{L}_{cls}$, box loss $\mathcal{L}_{box}$, and mask loss $\mathcal{L}_{mask}$. Among them, box loss (L1 loss $\mathcal{L}_{L1}$ and GIOU loss~\cite{rezatofighi2019generalized} $\mathcal{L}_{giou}$) and classification loss (focal loss~\cite{lin2018focal}) are the same as DINO~\cite{zhang2022dino}. For mask loss, we adopt cross-entropy $\mathcal{L}_{ce}$ and dice loss $\mathcal{L}_{dice}$. We also follow \cite{kirillov2020pointrend,cheng2021pointly,cheng2021mask2former} to use point loss in mask loss for efficiency. Therefore, the total loss is a linear combination of three kinds of losses: $\lambda_{cls}\mathcal{L}_{cls}+\lambda_{L1}\mathcal{L}_{L1}+\lambda_{giou}\mathcal{L}_{giou}+\lambda_{ce}\mathcal{L}_{ce}+\lambda_{dice}\mathcal{L}_{dice}$, where we set $\lambda_{cls}=4,\lambda_{L1}=5,  \lambda_{giou}=2,\lambda_{ce}=5$, and $\lambda_{dice}=5$.
\\\textbf{Basic hyper-parameters: }Mask DINO has the same architecture as DINO~\cite{zhang2022dino}, which is composed of a backbone, a Transformer encoder, and a Transformer decoder. Compared to DINO, we increase the number of decoder layers from six to nine and use $300$ queries. We follow Mask-RCNN~\cite{he2017mask} and Mask2Former~\cite{cheng2021mask2former} to setup the training and inference settings for segmentation tasks. We use batch size $16$ and train 50 epoch for COCO segmentation tasks (instance and panoptic), 160K iteration for ADE20K semantic segmentation, and $90K$ iterations for Cityscapes semantic segmentation.
We set the initial learning rate (lr) as $1\times 10^{-4}$ and adopt a simple lr scheduler, which drops lr by multiplying 0.1 at the 11-th epoch for the 12-epoch setting and the 20-th epoch for the 24-epoch setting. For the other segmentation settings, we drop the lr at 0.9 and 0.95 fractions of the total number of training steps by multiplying 0.1. Under the ResNet-50 backbone, we use $4$ A100 GPUs each with 40GB memory for all tasks. We report the frames-per-second (fps) tested on the same A100 NVIDIA GPU for Mask2Former and Mask DINO by taking the average computing time with batch size 1 on the entire validation set.
\\\textbf{Augmentations and Multi-scale setting: }We use the same training augmentations as in Mask2Former~\cite{cheng2021mask2former}, where the major difference from DINO~\cite{zhang2022dino} on COCO is that we use large-scale jittering (LSJ) augmentation~\cite{du2021simple,ghiasi2021simple} and a fixed size crop to $1024\times 1024$, which also works well for detection tasks. We use the same multi-scale setting as in DINO~\cite{zhang2022dino} to use 4 scales in ResNet-50-based models and 5 scales in SwinL-based models.
\subsection{Denoising training}\label{sec:denoise}
Following DN-DETR~\cite{li2022dn}, we train the model to reconstruct the ground-truth objects given the noised ones. These noised objects will be concatenated with the original decoder queries during training, but will be removed during inference. We add noise to both the bounding box and labels, which will serve as positional embedding and content embedding input to decoder queries. As a box can be viewed as a noised version of a segmentation mask, our unified denoising training will reconstruct the masks given the noised boxes, which improves segmentation training. 
\\\textbf{Label noise: } For label noise, we use \emph{label flip}, which randomly flips a ground-truth label into another possible label in the dataset with probability $p$. After adding noise, all the labels will go through a label embedding to construct high-dimensional vectors, which will be the content queries of the decoder. $p$ is set to 0.2 in our model. 
\\\textbf{Box noise: }
A box can be formulated as $(x, y, w, h)$, which is also the positional query of DINO~\cite{zhang2022dino}. We add two kinds of noise to the box including \emph{center shifting} and \emph{box scaling}. 
For center shifting, we sample a random perturbation $(\Delta x, \Delta y)$ to the box center. The sampled noise is constrained to $|\Delta x|<\frac{\lambda_1 w}{2}$ and $|\Delta y|<\frac{\lambda_1 h}{2}$, where $\lambda_1\in(0,1)$ is a hyperparameter to control the maximum shifting. For box scaling, the width and height of the box are randomly scaled to $\left[(1-\lambda_2), (1+\lambda_2)\right]$ of the original ones, where $\lambda_2$ is also a hyperparameter to control the scaling. In our model, we set $\lambda_1=\lambda_2=0.4$.

\section{Large models setting}
For large models with the SwinL backbone, we follow the same setting of DINO~\cite{zhang2022dino} to pre-train a model on the Objects365~\cite{shao2019objects365} dataset for object detection. Then we finetune the pre-trained model on COCO instance and panoptic segmentation for 24 epochs and on ADE20K semantic segmentation for 160k iterations. For training settings on instance and panoptic segmentation on COCO,  we use $1.2\times$ larger scale ($1280\times 1280$) and $16$ A100 GPUs. For training settings on ADE20K semantic, we use $3\times$ more queries ($900$) and $8$ A100 GPUs. We also use Exponential Moving Average (EMA) in this setting, which helps in ADE20K semantic segmentation.
\section{SOTA Results on COCO test-dev}
We show the COCO test-dev results in Table \ref{tab:testsota}.
\end{document}